\documentclass[journal]{IEEEtran}

\usepackage{cite}

\ifCLASSINFOpdf

\else

\fi

\usepackage{times}  % DO NOT CHANGE THIS
\usepackage{helvet} % DO NOT CHANGE THIS
\usepackage{courier}  % DO NOT CHANGE THIS
\usepackage[hyphens]{url}  % DO NOT CHANGE THIS
\usepackage{graphicx} % DO NOT CHANGE THIS
\usepackage{graphicx}  % DO NOT CHANGE THIS
\usepackage{amsmath,graphicx}

\usepackage{amsmath}
\usepackage{xcolor,soul,framed} %,caption
\usepackage{xspace}
\usepackage{array}
\usepackage{eqparbox}
\usepackage{url}
\usepackage{longtable}
\usepackage{lipsum}
\usepackage{blindtext}
\usepackage{makecell}
\usepackage{mathtools}
\usepackage{commath}
\usepackage{multirow}
\usepackage{tabularx}
\usepackage[normalem]{ulem}
\usepackage{xspace}
\usepackage{algorithm}
\usepackage{algpseudocode}
\usepackage{amsfonts}
\usepackage{placeins}

\newcolumntype{C}{>{\hsize=\dimexpr0.5\hsize+8\tabcolsep+\arrayrulewidth\centering\relax}X}
\DeclareMathOperator{\cref}{ref}

\usepackage{xcolor}
\usepackage{times}

\usepackage{soul}
\usepackage{url}
\usepackage[hidelinks]{hyperref}
\usepackage[utf8]{inputenc}
\usepackage[small]{caption}
\usepackage{graphicx}
\usepackage{amsmath}
\usepackage{booktabs}
\usepackage{amsfonts}
\usepackage{commath}
\usepackage{multirow}
\usepackage{tabularx}
\usepackage{mathtools}

\begin{document}

% \title{Learning EEG Representations on Riemannian Manifold}
% \title{Deep EEG Representation Learning with Riemannian Manifold}
% \title{Deep EEG Representation Learning on Riemannian Manifold for Affective Computing and Motor-Imagery Classification}
% \title{EEG-Based Deep Learning with Riemannian Manifold for Affective Computing and Motor-Imagery Classification}
% \title{Deep EEG Representation Fusion with Riemannian Manifold for Affective Computing and Motor-Imagery Classification}
% \title{Deep EEG Representation Fusion with Riemannian Geometry for BCI Applications}

\title{Distilling EEG Representations via Capsules for Affective Computing}% in Smart Vehicles}

% \title{Multimodal EEG-EOG Representation Learning with Deep Attentive Capsules for Driver\\Vigilance Estimation}% in Smart Vehicles}

% \title{Monitoring EEG and EOG with Attentive Capsules for Driver Vigilance Estimation}% in Smart Vehicles}
% \title{Learning EEG and EOG with Attentive Capsules for Smart Vehicle Applications}
% \title{Deep Attention for Human Attention:\\Driver EEG and EOG Monitoring with Attentive Capsules for Vigilance Estimation in Smart Vehicles}
% \title{Driver Monitoring with Multimodal EEG-EOG Learning using a Deep Recurrent Network and Capsule Attention}
% \title{Driver Monitoring using Deep Recurrent Multimodal EEG-EOG Representation Learning with Capsule Attention}
% \title{Driver Monitoring using Deep Recurrent Multimodal EEG-EOG Spatiotemporal Representation Learning with Capsule Attention}
% \title{Capsule Attention for Multimodal EEG and EOG Spatiotemporal Representation Learning with Application to Driver Monitoring}
% \title{Attention-for-Attention: Deep Recurrent Neural Network with Attentive Capsule Routing for Multimodal Driver Vigilance Estimation}
% Capsule Attention for Multimodal EEG and EOG Spatiotemporal Representation Learning with Application to Driver Monitoring
% Attention-for-Attention: Deep Recurrent Neural Network with Attentive Capsule Routing for Multimodal Driver Vigilance Estimation
% Driver Monitoring with Multimodal EEG-EOG Learning using a Deep Recurrent Network and Capsule Attention

\author{Guangyi~Zhang 
        and~Ali~Etemad,~\IEEEmembership{Senior Member,~IEEE}% <-this % stops a space
\thanks{G. Zhang, and A. Etemad are with the Department of Electrical and Computer Engineering, Queen's University, Kingston, ON, Canada (e-mail: guangyi.zhang@queensu.ca,ali.etemad@queensu.ca).}

}

\maketitle

\begin{abstract}
Affective computing with Electroencephalogram (EEG) is a challenging task that requires cumbersome models to effectively learn the information contained in large-scale EEG signals, causing difficulties for real-time smart-device deployment. In this paper, we propose a novel knowledge distillation pipeline to distill EEG representations via capsule-based architectures for both classification and regression tasks. Our goal is to distill information from a heavy model to a lightweight model for subject-specific tasks. To this end, we first pre-train a large model (teacher network) on large number of training samples. Then, we employ the teacher network to learn the discriminative features embedded in capsules by adopting a lightweight model (student network) to mimic the teacher using the privileged knowledge. Such privileged information learned by the teacher contain similarities among capsules and are only available during the training stage of the student network. We evaluate the proposed architecture on two large-scale public EEG datasets, showing that our framework consistently enables student networks with different compression ratios to effectively learn from the teacher, even when provided with limited training samples. Lastly, our method achieves state-of-the-art results on one of the two datasets. %We make our implementation publicly available: \textcolor{magenta}{\href{run:URL withheld for anonymity}{URL withheld for anonymity}}.
% Especially, the largest improvement is observed on the student with the smallest parameters. 
% helps the student networks on performance when the compression ratio changes. 
% the competitive performance achieved by the lightweight network with our novel pipeline.
% Then, we employ the teacher network to learn the discriminative features embedded in capsules from intra-subject data. Next, we adopt a lightweight model (student network) to mimic the teacher using the privileged knowledge. Such privileged information learned by the teacher contain similarities among capsules and are only available during the training stage of the student network. We evaluate the proposed architecture on two large-scale public EEG datasets, showing the competitive performance achieved by the lightweight network with our novel pipeline.  We make our implementation publicly available: \textcolor{magenta}{\href{run:URL withheld for anonymity}{URL withheld for anonymity}}.
\end{abstract}

\section{Introduction}

% Edited
Affective computing is a field of research concerned with building computational models for emotion recognition in order to help computers understand, analyze, and mimic human emotions \cite{picard2000affective}. Non-invasive technologies such as recording of brain signals with Electroencephalogram (EEG) have been widely used for affective computing \cite{zheng2015investigating,zheng2017multimodal,zhang2020rfnet}. 

% Edited
EEG is a non-stationary time-series, generally with a large number of dimensions (high dimensionality) and sampling rate (temporal resolution). As a result, recent deep learning solutions for EEG representation learning often require complex networks to sufficiently learn the information contained within EEG signals. Specifically, Capsule Networks (CapsNet) \cite{sabour2017dynamic} have been applied to EEG for affective computing and achieved state-of-the-art results \cite{zhang2019capsule}. Nonetheless, such approaches that rely on capsules for EEG representation learning consist of a large number of parameters, making it difficult for online and real-time use for smart device deployment.

% Edited
In this paper, we propose a novel method for knowledge distillation based on capsule networks, capable of being used for both classification and regression tasks in the context of EEG representation learning. Specifically, we first revisit the capsule-based model proposed in \cite{zhang2019capsule} and use a similar architecture as our student network.
% teacher-student network, where The student model is smaller than teacher. The higher level capsules in the architecture can be used to provide soft target labels, thus enabling the distillation for both classification and regression tasks. 
% We then develop a novel knowledge distillation mechanism via capsules. Our distillation mechanism not only focuses on similarities among the global information embedded in higher level capsules, but also the local features contained in their lower level capsules. 
We then develop a novel knowledge distillation framework via capsules, which transfers knowledge contained in both higher and lower level capsules from the teacher to the student. Next, in order to utilize more training data, we pre-train the teacher network on large amounts of cross-subject EEG data. We then fine-tune the pre-trained teacher on intra-subject data to learn subject-specific knowledge.
% due to the biological discrepancy among subjects. 
Afterwards, we evaluate the student on subject-specific experiments with the help of privileged information learned by the teacher and transferred to the student. 
% Overall, we transfer the privileged information learned by teacher to student via our novel knowledge distillation process. 
Our experiments show that a compact student network with only $0.03\%$ of the number of parameters of the original teacher network can achieve competitive results in comparison. 
% Moreover, our approach enables training to be performed with fewer data without loss of performance, and in fact in many cases with an improvement in performance. 
Moreover, our approach improves the robustness of the compact student when faced with limited training samples.
Lastly, our experiments on two separate public datasets show competitive results by the student model for one of the datasets, while outperforming the related work on the other dataset.

% We use the similar Capsule Attention (CapsAtt) based Long Short-term Memory (LSTM) architecture that successfully implemented in vigilance estimation as our teacher model \cite{zhang2019capsule}. We investigate the knowledge distillation using both  local information contained in lower level capsules and global information in higher level capsules \cite{sabour2017dynamic,zhang2019capsule}. 

% Edited
Our contributions are summarized as follows. $\textbf{(1)}$ For the first time, we propose a distillation pipeline based on a teacher-student framework capable of capsule network compression, while improving overall student performance. $\textbf{(2)}$ To the best of our knowledge, this is the first time that knowledge distillation has been implemented for affective EEG representation learning. $\textbf{(3)}$ Our proposed method can be used for both classification and regression tasks. Experiments on SEED and SEED-VIG datasets show that our model performs well on SEED, while achieving a \textit{new state-of-the-art} on SEED-VIG.
%(4) We make out implementation public...
% $\textbf{(4)}$ We make our source code for conducting experiments publicly available at \footnote{URL withheld for anonymity} to enable reproducibility.
%$\textbf{(4)}$ We make the source code publicly available~\footnote{URL withheld for anonymity} for reproducibility purposes and contributing to the field.

% We investigate distillation in capsule level manner by revisiting the LSTM-CapsAtt network. 

\section{Related Work}
% Edited
\textbf{Affective Computing with EEG.} A variety of deep learning techniques have been used to learn the most discriminative features extracted from EEG for affective computing. For example, 
% Zheng \textit{et al.} utilized a Deep Belief Network (DBN) to extract higher level features by implementing multiple deep hidden layers for emotion recognition \cite{zheng2015investigating}. 
In \cite{wu2018regression}, the authors adopted Double-Layered Neural Network with Subnetwork Nodes (DNNSN) to estimate drivers' vigilance levels . In  \cite{huo2016driving}, a Graph regularized Extreme Learning Machine (GELM) was employed to predict fatigue . To learn the time-dependency and spatial information in EEG signals, the authors used Spatial-Temporal Recurrent Neural Network (STRNN) \cite{zhang2018spatial}, achieving strong performance. In addition to spatiotemporal feature learning, Regional to Global Brain-Spatial-Temporal Neural Network (R2G-STNN) was proposed to minimize the domain-shift by applying a discriminator \cite{li2019regional}. To further investigate dependencies between EEG electrodes, in \cite{zhang2020variational}, the authors proposed Variational Pathway Reasoning (VPR) for emotion classification, achieving state-of-the-art results. The VPR pipeline employed Long short-term memory (LSTM) to learn sequential information between electrodes, thus encoding the pathway around them. Then this method used a Bayesian probabilistic approach to learn pathways' scaling factors to identify the one with the most salient pair-wise connections \cite{zhang2020variational}.  In \cite{zhong2020eeg}, Regularized Graph Neural Networks (RGNN) was employed to to explore the graph connections of EEG electrodes, approaching the best results with fully explored topological knowledge. In \cite{zhang2019capsule}, an LSTM-CapsNet model was proposed to explore both temporal and spatial information to predict vigilance. Very Recently, Zhang and Etemad proposed a Riemannian Fusion Network (RFNet) to learn the temporal information through an LSTM-attention network \cite{zhang2020rfnet}. This method also learned spatial information through a parallel Riemannian-based approach with Spatial Covariance Matrix (SCM) as input.

% Edited
\textbf{Capsule Networks.} Capsule networks were proposed in \cite{sabour2017dynamic} to learn the \textit{part-whole} relationships of objects through iterative routing among different level capsules. The pipelines using capsule networks have achieved state-of-the-art results in some areas of Natural Language Processing (NLP) such as intent detection \cite{zhang2018joint} and multi-label classification \cite{chen2020hyperbolic}, as well as computer vision such as expression recognition \cite{sepas2101capsfield} and low resolution image recognition \cite{singh2019dual}. Recently, an LSTM-CapsNet architecture was successfully proposed for EEG-based affective computing \cite{zhang2019capsule}. 

% Edited
\textbf{Knowledge Distillation.}
% Prior to knowledge the idea of distillation by Hinton \textit{et al.} \cite{hinton2015distilling}, 
Vapnik \textit{et al.}  proposed a learning paradigm to enable machine learning from other machines with privileged information in the training stage \cite{vapnik2009new,vapnik2015learning}. The paradigm relies on a teacher machine with more discriminative information than the student machine, using Support Vector Machine (SVM). The teacher machine is effective when its expected error is smaller than the student's, as theoretically shown in \cite{vapnik1998statistical}. However, there are several constraints in the paradigm such as the restriction to SVM, fixed parameters, and lack of information on hard labels in the student machine training \cite{lopez2015unifying}. 

% Edited
In order to develop an effective knowledge transfer pipeline suitable for deep learning, knowledge distillation has been proposed in \cite{hinton2015distilling}. Knowledge distillation transfers soft target distributions learned by a cumbersome model (teacher) to a smaller model (student), where the architecture and parameters of the models can be customized \cite{hinton2015distilling}. For this purpose, KL divergence of the soft target distribution is minimized between the student and teacher networks, thus enabling \textit{pure} knowledge distillation during the training stage \cite{hinton2015distilling}. In \cite{lopez2015unifying}, knowledge distillation was described as a two-step process in which \textit{i}): the teacher learns the data using hard labels; \textit{ii}): the student learns the data using soft labels computed from the teacher, as well as the hard labels. Consequently, the student not only receives the privileged knowledge from the teacher, but also learns the hard labels during the distillation process. In \cite{phuong2019towards}, analysis of data geometry and optimization bias also showed the benefits of knowledge distillation. Very recently, knowledge distillation has been successfully implemented in a number of computer vision areas such as video captioning \cite{Pan_2020_CVPR} and prediction regularization \cite{Yun_2020_CVPR}. This concept has also been recently used in different areas of NLP such as large model compression \cite{sun2019patient} and natural language generation \cite{tang2019natural}.

% Edited
\section{Our Approach}
\textbf{Overview.} We aim to distill information from a heavy and cumbersome capsule-based model to a lightweight model for subject-specific tasks suitable for both classification and regression. To do so, we propose a four-step process. (\textit{i}) Developing two separate networks containing CapsNet architectures, one called the \textit{teacher} network and the other the \textit{student} network. (\textit{ii}) Pre-training the teacher network on the large amounts of available cross-subject data. (\textit{iii}) Using the pre-trained teacher to then learn information embedded in capsules with intra-subject data. (\textit{iv}) Training the student on intra-subject data with the help of the privileged information learned by the teacher via capsules. As shown by our results in Section 4, this process enables us to \textbf{maximally} compress the model with minimal loss in performance.

% Our goal is to transfer knowledge from a cumbersome model (teacher) to a lightweight network (student) with a generalized pipeline for both classification and regression tasks. 

% Edited
In the following sections, we first revisit the architecture of the LSTM-CapsNet model used in this study, and then introduce our novel distillation framework via capsules.

% Edited
\subsection{Revisiting LSTM-CapsNet Architecture}
\textbf{Feature Space.} The input EEG data are first pre-processed followed by feature extraction. Pre-processing of EEG has been kept consistent with previous works on the same datasets \cite{zheng2015investigating,zheng2017multimodal}. Specifically, the signals were downsampled to $200$ Hz from $1000$ Hz. A band-pass filter of $[0.5-70]$ Hz followed by a notch filter at $50$ Hz were then applied to the raw EEG signals to minimize artifacts and scale down power line noise. Data normalization was followed to scale signal amplitudes into the range of $[-1, 1]$, thus reducing the discrepancy of EEG collected from various subjects and data recording periods \cite{zhang2020rfnet}. 

% Edited
% Prior to feature extraction,
In the feature extraction step, we extracted two types of features notably Power Spectrum Density (PSD) and Differential Entropy (DE). To extract PSD features, firstly, we applied consecutive $1$-second Hanning windows with no overlap on each $L$-second EEG segment, thus avoiding spectral leakage caused by finite windowing (the value for $L$ along with other parameters used for pre-processing and the network architectures, are presented later in Section 4). We then have a total of $L$ number of Hanning windows in each EEG segment. We then applied Short-Time Fourier Transform with the Hanning window to transform signals from time domain to frequency domain. Afterwards, we compute the logarithm of PSD using Eq. \ref{equation: psd} and DE using Eq. \ref{equation: de} (with the assumption of the Gaussian distribution of the signal) in different frequency bands \cite{zhang2020rfnet}.
% To extract DE features, we first applied a filter bank to filter signals from each $1$-second window into the same several frequency bands in PSD calculation. 
% Next, we compute  in each frequency band \cite{zhang2020rfnet}. 
\begin{equation}\label{equation: psd}
S_{xx}(\omega)=\lim_{t\to\infty}E\Big[|\hat{X}(\omega)|^{2}\Big].
\end{equation}
% \vspace{-3mm}
\begin{equation}\label{equation: de}
DE = \frac{1}{2} \log{2\pi e \sigma^{2}}, \hspace{3mm} if x\sim \mathcal{N}(\mu, \sigma^{2}).  
\end{equation}

% Edited
\textbf{LSTM Network.} 
% LSTM proposed by \cite{hochreiter1997long} is one type of RNN capable of addressing gradients vanish or exploding challenges while handling long-term dependencies. LSTM have been successfully implemented in sequential data (e.g., EEG) for time-dependent tasks (e.g., vigilance estimation) \cite{zhang2019classification,zhang2019capsule,zhang2020rfnet}. 
We employ an LSTM network to learn the time-dependencies within the EEG signals. Specifically, we feed the extracted features from each of the $L$ number of windows to the corresponding $L$ number of cells of the input LSTM layer with $M$ hidden units. The time-dependent information learned by each LSTM cell is reshaped from $M$ units to a $[\sqrt M, \sqrt M]$ square matrix for further processing.

% Edited
\textbf{Lower Level Capsules.} We employ a 2D convolution layer with $L$ number of output channels, kernel size of $3$, and stride of $1$, to capture local features. We then apply another 2D convolution layer with $C$ output channels, $1$ kernel and $1$ stride to produce lower level capsules, yielding $A =[C\times (\sqrt M -2)^2]$ number of lower level capsules with $d = (L/C)$ dimensions.

% Edited
\textbf{Higher Level Capsules.} Higher level capsules are designed to learn global information as opposed to lower level capsules which capture local information \cite{sabour2017dynamic}. Let's denote $K$ and $H$ as the number and dimension of higher level capsules, respectively. $K$ is consistent with the number of categories in our classification task and can be empirically tuned for regression tasks. We enable the higher level capsules to have larger degrees of freedom by setting $H > d$ \cite{sabour2017dynamic}.

% Edited
\textbf{Capsule Network.} Capsule networks have been used to learn the `part-whole' relationships between lower level capsules and higher level capsules \cite{sabour2017dynamic}. 
% As shown in Figure \ref{capsule attention},
The capsule network assigns attention scores from lower level capsules ('part' information) to higher level ones ('whole' information) through dynamic routing \cite{sabour2017dynamic}. Specifically, a CapsNet used as an attention mechanism is established between the prediction vector $\hat u_{j|i}$ and the output of higher level capsules ($v_j$). $\hat u_{j|i}$ represents the prediction from each lower level capsule $i \in [1, A]$ to each higher level capsule $j \in [1, K]$. The prediction vector is expressed as the multiplication of weight matrix $W_{ij} \in \mathbb{R}^{d\times H}$ with the output of lower level capsule $u_i$. $s_j = \sum_i c_{ij}\hat{u}_{j|i}$ represents the total input to capsule $j$. The output of higher level capsule $v_j$ is the squashed output of $s_j$ which normalizes $v_j$ into the range of $(0,1)$. $c_{ij}$ are the softmax outputs of logit $b_{ij}$, where $b_{ij}$ represents the log prior probabilities that will be updated by the iterative process $b_{ij} \leftarrow b_{ij} + \hat u_{j|i}\cdot v_j$ after its zero initialization.
% , as illustrated in Figure \ref{capsule attention}. 

\begin{figure}[!t]
    \begin{center}
    \includegraphics[width=1.0\linewidth]{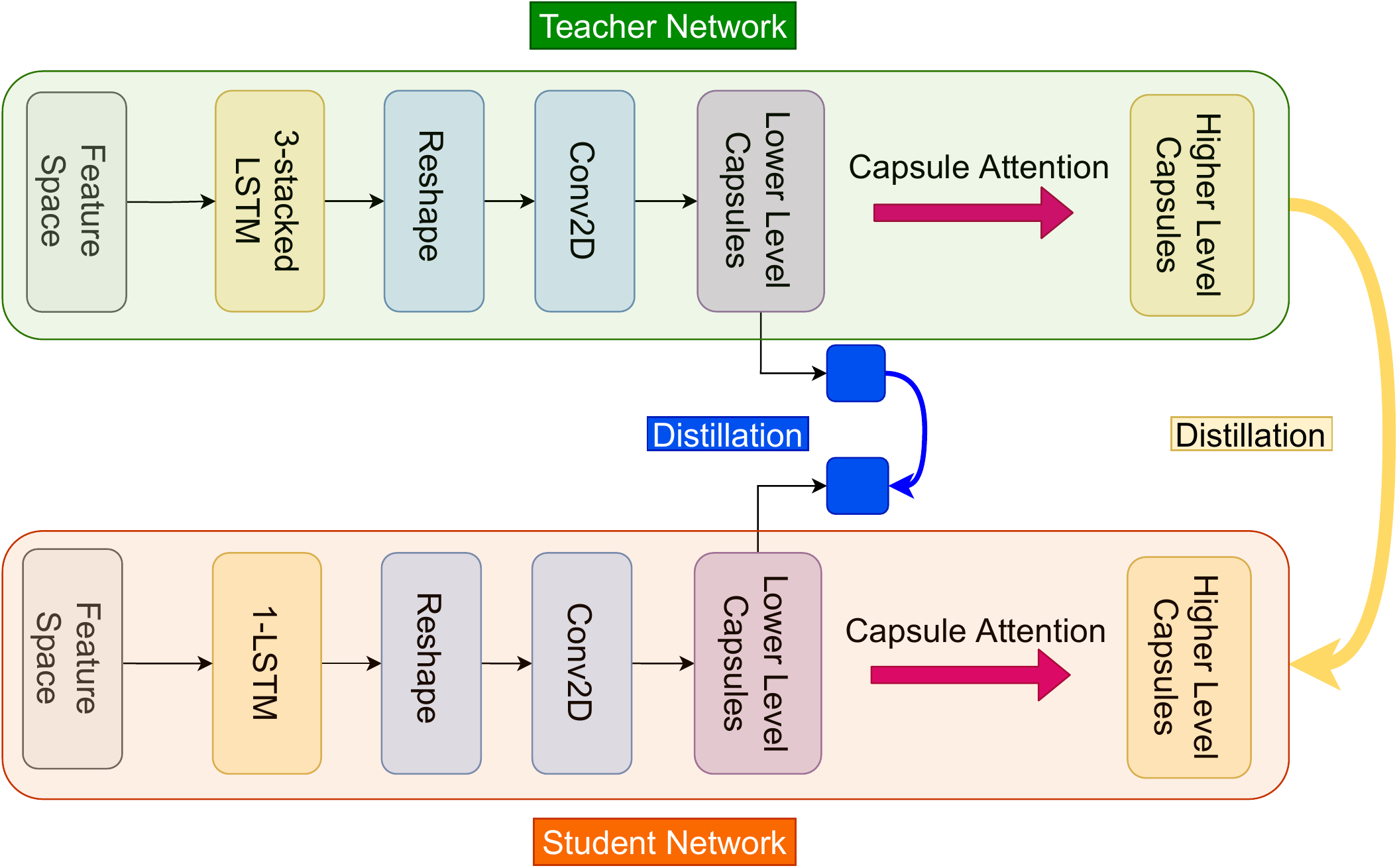}
    \end{center}    
\caption{The overview of our novel knowledge distillation framework is presented.}
\label{fig-overview}
\end{figure}

% \begin{figure}[!t]
%     \begin{center}
%     \includegraphics[width=0.5\linewidth]{3-2.pdf} 
%     \end{center}
% \caption{Capsule Attention}
% \label{capsule attention}
% \end{figure}

% Edited
\subsection{Proposed Method for Knowledge Distillation}
An overview of our novel knowledge distillation framework is illustrated in Figure \ref{fig-overview}. 
% Our goal is to transfer knowledge from a cumbersome model (teacher) to a lightweight network (student) with a generalized pipeline for both classification and regression tasks. 
We develop a knowledge distillation framework to compress the large model without performance degradation. To do so, we first employ the LSTM-CapsNet architecture described above as the teacher network. Next, we pre-train the teacher on cross-subject data and then fine-tune it on intra-subject data in order to adapt to subject-specific features. At last, we train the student model with the help of privileged information learned by the teacher, and then evaluate it on intra-subject data. In order to fully explore the privileged information, we learn the inter-dimension relationships of lower level capsules, as well as the information embedded in higher level capsules, through minimizing their similarities between the teacher and student networks. 

% we design a loss function 

% and employ the smaller version of the teacher network as student networks. 
% We pretrain the teacher on larger dataset and train the student on smaller dataset. Prior to knowledge distillation, we \textbf{fine-tune} the pre-trained teacher network on smaller dataset for student work training. During the fine-tuning, we freeze the LSTM architecture to take advantage of features extracted and learned from pre-training in larger dataset. We unfreeze the parameters in both capsule layers, in order to adapt the same features that student network uses in the smaller dataset.  

% Edited
\textbf{Knowledge Distillation via Lower Level Capsules.} Lower level capsules contain local features where their different dimensions capture different aspects of the information space \cite{sabour2017dynamic,zhang2019capsule}. For example, when trained with handwritten digital images, each dimension of lower level capsules contains different information on digital-specific variations, for instance, scale, thickness, and width \cite{sabour2017dynamic}. Therefore, we explore the similarity \cite{tung2019similarity} of the capsules' inter-dimension correlations between the teacher and student networks. In order to learn such inter-dimension correlations, we first calculate the covariance matrix of the lower level capsules as:
\begin{equation}
    G = u^T \cdot u, u \in \mathbb{R}^{A\times d} ; G' = u'^T \cdot u', u' \in \mathbb{R}^{A'\times d}, 
\end{equation} 
where $G, G' \in \mathbb{R}^{d\times d}$ represent the covariance matrices of lower level capsules of the teacher ($u$) and student ($u'$) networks respectively. Dimension $d$ is kept the same for both networks. The number of lower level capsules of the teacher are greater than or equal to the student's as $A \geq A'$. Next, we compute the square euclidean distance between two $L2$ normalized covariance matrices \cite{tung2019similarity} as similarity loss:
\begin{equation}
    \mathcal{L}_{U} = ||\frac{G}{||G||_2} - \frac{G'}{||G'||_2} ||_F^{2},
\end{equation} 
where $||.||_F$ is the Frobenius norm.

% Edited
\textbf{Knowledge Distillation via Higher Level Capsules.} Higher level capsules include global information, where the length of their output vectors $v_j$ represents the probability that the entity corresponding to that capsule $j$ exists. Such information is further used as 'soft target labels' for knowledge distillation. 
% We then focus on the similarities among higher level capsules.
Specifically, we employ KL divergence to measure the difference of information distribution in higher level capsules between teacher ($v_j$) and student network ($v_j'$). The following equation is used:
\begin{equation}
    \mathcal{L}_{V} = KL[\log (\sigma (||v_j'||/\tau)), \sigma (||v_j||/\tau)]\tau^2 ,
\end{equation}
where $\sigma (.)$ is the softmax operator, $\tau$ is the temperature parameter \cite{hinton2015distilling}, and a logarithm operation is applied to the student output to help accelerate the distillation process. We use $\tau=1$ throughout the experiments.

% Edited
\textbf{Teacher Network.} We use similar architecture to the state-of-the-art network from \cite{zhang2019capsule} as our teacher network. We employ $N$-stacked LSTM layers with $256$ units. Layer normalization \cite{ba2016layer} is used and followed by LeakyReLU after each LSTM layer. 
% Other parameters are kept the same as in \cite{zhang2019capsule}. 
For the SEED dataset, we set the number of higher level capsules to $K=3$ to be consistent with the number of emotion classes, in order to use margin loss for classification \cite{sabour2017dynamic}. For the regression task in the SEED-VIG dataset, we empirically set $K=10$ as the number of higher level capsules. The output is then followed by a fully connected layer containing $10$ hidden units with Sigmoid activation. The details are shown in Table \ref{table:implemenation}.

% Edited
\textbf{Student Network.} The student network has the same architecture as the teacher but with fewer parameters. Specifically, the student network contains a single layer of LSTM with fewer hidden units $M \leq 256$, yielding a smaller number of lower level capsules $A$. The parameter details are presented later in Section 4.4. 

% Edited
\textbf{Training Loss Function.} The training loss function includes three parts, namely lower level capsule distillation loss, higher level capsule distillation loss, as well as a task-specific loss. The task-specific loss depends on the task type (classification vs. regression). For the regression task, we employ a fully connected layer ($K=10$ hidden units) with a sigmoid activation function as in \cite{zhang2019capsule} to enable the Minimum Squared Error (MSE) loss calculation ($\mathcal{L}_{MSE}$). For the classification task, we use margin loss (Eq. \ref{margin loss}) as recommended in \cite{sabour2017dynamic}: 
\begin{equation}\label{margin loss}
    \begin{split} 
    \mathcal{L}_{k} =& T_k \max(0, 0.9-||v_k||)^2 + \\ &0.5 \times (1-T_k) \max(0, ||v_k||-0.1)^2,
    \end{split}
\end{equation} 

where $T_k=1$ if class $k$ is the correct prediction, otherwise $T_k=0$. The first part of the equation will be zero if and only if the probability of correct prediction is greater than $0.9$. The second part of the loss function will be zero if and only if the probability of incorrect prediction is less than $0.1$. 

% Edited
Consequently, the total loss is shown as: 
\begin{equation}\label{total loss}
    \mathcal{L}_{total} =  \eta \xi \mathcal{L}_{U} + \alpha\mathcal{L}_{V} + \begin{cases}
    (1-\alpha)\mathcal{L}_k,& \text{if classification}\\
    (1-\alpha)\mathcal{L}_{MSE},& \text{if regression}
\end{cases}
\end{equation}
where $\xi$ is the scaling factor, and $\eta$ and $\alpha$ are trade-off hyper-parameters for lower and higher level capsules distillation loss, respectively.

\begin{table*}[!ht]
\centering
\small
\setlength\tabcolsep{8.0pt}
\caption{Implementation details of the teacher network.}\label{table:implemenation}
\begin{tabularx}{1.9\columnwidth}{c|c|c|c|c|c|c|c|c|c}
  
    \hline
          & {Feature Space} &\multicolumn{2}{c|}{Temporal Info.} & \multicolumn{3}{c|}{Lower Level Capsules} & \multicolumn{2}{c|}{Higher Level Capsules} &{Regression Layer} \\
     \hline
     Dataset       & Features No.     & $L$ & $M$       & $C$ & $A$    & $d$    & $K$ & $H$ & Activation\\ %(W.O Dimension Reduction)
     \hline \hline
     SEED          & $620$           & $8$    & $256$  & $2$ & $392$  & $4$    & $3$ & $16$ & N/A\\
     SEED-VIG      & $850$           & $8$    & $256$  & $2$ & $392$  & $4$    & $10$& $16$ & Sigmoid\\

	\hline
% 	\hline
% 	\hline
\end{tabularx}
\end{table*}

\begin{table*}[!t]
\centering
% \scriptsize
\caption{Comparison of different solutions and results for the SEED dataset.}\label{table:SEED}
\setlength\tabcolsep{10.0pt}
\begin{tabularx}{0.5\textwidth}{lllc}
    \hline
     Paper  & Input & Method & Acc.$\pm$SD \\
	\hline
	\hline

     \cite{zhang2018spatial}              & DE              & STRNN                    & $0.8950\pm 0.0763$ \\
     \cite{li2019regional}                & DE              & R2G-STNN                 & $0.9338\pm 0.0596$ \\
     \cite{zhang2020rfnet}                &  SCM,DE, PSD    & RFNet                    & $0.9372\pm 0.0571$ \\
     \cite{zhong2020eeg}                  & DE              & RGNN                     & $0.9424\pm 0.0595$ \\
     \cite{zhang2020variational}          & DE              & VPR                      & $0.9430\pm 0.0650$ \\
    \hline
	\textbf{Ours}                         & DE, PSD         & Distillation             & $0.9107\pm 0.0763$ \\
	\hline
\end{tabularx}
\end{table*}

\begin{table*}[!t]
\centering
% \scriptsize
\caption{Comparison of different solutions and results for the SEED-VIG dataset.}\label{table:SEED-VIG}
\setlength\tabcolsep{10.0pt}
\begin{tabularx}{0.7\textwidth}{lllcc}
    \hline
     Paper & Input  & Method & RMSE$\pm$SD & PCC$\pm$SD\\
	\hline
	\hline
	 \cite{huo2016driving}                   & DE           & GELM                   & $0.1037\pm 0.0309$ & $0.7013\pm 0.1045$ \\	
	 \cite{wu2018regression}                 & DE           & DNNSN                  & $0.1175\pm 0.0420$ & $0.7201\pm 0.1706$ \\
	 \cite{zhang2019capsule}                 & DE, PSD      & LSTM-CapsNet           & $0.0295\pm 0.0095$ & $0.9887\pm 0.0072$ \\
	 \cite{zhang2020rfnet}                   & SCM, DE, PSD & RFNet                  & $0.0348\pm 0.0265$ & $0.9890\pm 0.0081$ \\
	 
	\hline
	\textbf{Ours}                            & DE, PSD      & Distillation       & $\mathbf{0.0258\pm 0.0095}$ & $\mathbf{0.9930\pm0.0047}$ \\
	\hline
\end{tabularx}
\end{table*}

\begin{table}[!t]
\centering
\caption{Details of the student network with different sizes.}\label{table:Distillation_Details}
\setlength\tabcolsep{6.0pt}
\begin{tabularx}{1.0\columnwidth}{lccccc}
    \hline
    
     Model              & $N$        & $M$          & $A$        & Params.    & Compress. Ratio\\
	\hline\hline
	Teacher             & $3$        & $256$        & $392$       & $1.50$M & $1.00$       \\	
	\hline
	Student $1$         & $1$        & $256$        & $392$       & $0.97$M  & $0.64$  \\	
% 	Student $2$         & $1$        & $196$        & $288$       & $0.70$M  & $0.47$  \\	
	Student $2$         & $1$        & $144$        & $200$       & $0.48$M   & $0.32$  \\
	Student $3$         & $1$        & $64$         & $72$        & $0.19$M   & $0.13$  \\
	Student $4$         & $1$        & $16$         & $8$         & $0.04$M   & $0.03$  \\

%%%%%%%%%%%%%%%%%%%%%%%%%%%%%%%
% What is the correct compression ratio??
	\hline
\end{tabularx}
\end{table}

% \begin{table}[!ht]
% \centering
% \caption{Performance SEED}\label{table:model_size_seed}
% \setlength\tabcolsep{10.0pt}
% \begin{tabularx}{1.0\columnwidth}{lcc}
%     \hline
    
%      Model &Acc.&  Distill. Acc       \\
% 	\hline
% 	\hline
% 	Student $1$    &$0.9057\pm 0.0775$      &$0.9107\pm 0.0763$        \\	
% 	Student $2$    &$0.9001\pm 0.0779$      &$0.9070\pm 0.0770$         \\
% 	Student $3$    &$0.8891\pm 0.0780$      &$0.9033\pm 0.8289$         \\
% 	Student $4$    &$0.8831\pm 0.0915$      &$0.9021\pm 0.0803$          \\

% 	\hline
% \end{tabularx}
% \end{table}

% \begin{table*}[!ht]
% \centering
% \caption{Performance SEED-VIG}\label{table:model_size_seed-vig}
% \setlength\tabcolsep{10.0pt}
% \begin{tabularx}{2.0\columnwidth}{lcccc}
%     \hline
    
%      Model &RMSE.&  PCC & Distill RMSE .&  Distill PCC\\
% 	\hline
% 	\hline
% 	Student $1$         &$1$    & $256$ & $256$    \\	
% 	Student $2$         &$1$    & $256$ & $256$      \\
% 	Student $3$         &$1$    & $256$ & $256$     \\
% 	Student $4$         &$1$    & $256$ & $256$     \\

% 	\hline
% \end{tabularx}
% \end{table*}

% Edited
\section{Experiments}
\subsection{Datasets}
\textbf{SEED.} The SEED dataset was collected by \cite{zheng2015investigating} to perform emotion recognition with three categories of positive, negative, and neutral emotions. $15$ emotion-related videos were selected as stimuli in each experiment. $15$ subjects, including $8$ females and $7$ males, performed a total of $30$ experiments, where each subject participated in the experiments in two different runs. Each run contained $15$ sessions. Each session started with a $5$-second notice before playing the video clips, followed by approximately $4$ minutes of watching the movie clip, and concluded by $45$ seconds of self-assessment. Each session ended with a $15$-second relaxation. $62$ EEG channels were recorded with a sampling rate of $1000$ Hz using the international $10-20$ system. 
% EEG recordings are split into segments of $8$ seconds while no overlap exists between segments.

% Edited
\textbf{SEED-VIG.} The SEED-VIG dataset \cite{zheng2017multimodal} contains EEG recordings to estimate drivers' continuous vigilance levels. $23$ subjects ($12$ female and $10$ male) participated in the experiment and drove a simulated vehicle in a virtual environment. The experiment took around $120$ minutes. $885$ overall consecutive EEG segments were recorded in each experiment. The duration of eye blinks and eye closures as well as the duration of fixation and saccade \cite{zheng2017multimodal}, which were all measured using eye-tracking glasses, were used to measure the output ground-truth labels called PERCLOS.
% PERCLOS was used to annotate vigilance detection \cite{dinges1998perclos}. Specifically, PERCLOS was measured by eye movements captured by eye-tracking glasses. It is calculated by the duration of blinks and eyes closures (CLOS) over itself as well as the duration of fixation and saccade \cite{zheng2017multimodal}. 
The EEG signals were recorded from $17$ locations with a sampling rate of $1000$ Hz using the international $10-20$ system.

% Edited
\subsection{Evaluation Protocol}
\textbf{Teacher Network.} 
We pre-train the teacher network on the \textbf{cross-subject} data. We use leave-one-subject-out cross-validation to pre-train a teacher for each subject. Consequently, the pre-trained teacher used for each specific subject has not seen the data from that subject during training. 
% In the meantime, we are able to utilize the EEG features from other subjects that learned in the pre-trained model. 
For the SEED dataset, we have $418$ EEG trials for each experiment run per subject, yielding a total of $418\times2\times14 = 11704$ EEG trials for training, and $418\times2\times1 = 836$ EEG trials for testing. Similarly, in the SEED-VIG dataset, we have a total of $885\times22 = 19470$ EEG trials for training and $885$ EEG trials for testing.

% Edited
\textbf{Student Network.} 
We train and evaluate the student network on \textbf{intra-subject} data. We follow the same evaluation protocol as the related works \cite{zheng2015investigating,zheng2017multimodal,zhang2020rfnet}. In the SEED dataset, we use the pre-defined first $9$ sessions and the last $6$ sessions as the training set ($248$ EEG trials) and test set ($170$ EEG trials), respectively \cite{zheng2015investigating}. In the SEED-VIG dataset, we employ $5$-fold cross-validation for our train-test set split as in \cite{zheng2017multimodal}.

% Edited
\textbf{Evaluation Metrics.}
We adopt both Pearson Correlation Coefficient (PCC) and Root Mean Squared Error (RMSE) as evaluation metrics for the regression task in the SEED-VIG dataset \cite{zheng2017multimodal}, while accuracy (Acc.) is used as the evaluation metric for classification in the SEED dataset \cite{zheng2015investigating}.

% Edited
\subsection{Implementation Details}
\textbf{Feature Extraction.} We use different frequency bands in the feature extraction step for each dataset. For the SEED dataset, we use five frequency bands, notably delta, theta, alpha, beta, and gamma bands \cite{zheng2015investigating}. Accordingly we have $5 \times 2 \times 62 =620$ features extracted from each $1$-second window. For the SEED-VIG dataset, we use $25$ frequency bands with $2 Hz$ resolution, starting from $0.5$ to $50.5 Hz$ \cite{zheng2017multimodal}. We thus have $25 \times 2 \times 17 = 850$ features extracted from each window. 
% These details are summarized in Table \ref{table:implemenation}. 

% Edited
\textbf{Other Hyper-Parameters and Training.} In this work, we apply weight clipping to avoid gradient explosion. In the teacher pre-training phase, we run a total of $200$ epochs. The learning rate is initialized to $0.001$ and decreases by $10$ times after $100$th epoch, then drops again by $5$ times after the $150$th epoch. For the rest of the experiments (fine-tuning and subject-specific phases), training is performed with $50$ epochs, with a fixed learning rate of $0.001$. We employ the Adam algorithm with default decay rates for optimization. The batch sizes are set to $64$ during teacher pre-training and $8$ for all the other experiments. We set the scaling factor $\xi$ to $10^3$, and the trade-off hyper-parameter $\alpha$ to $0.7$ for SEED and $0.3$ for SEED-VIG, respectively. The parameter $\eta$ is set to be $0.3$ for both datasets. All hyper-parameters were empirically tuned on the validation set. All of our experiments are implemented using PyTorch \cite{paszke2019pytorch} on a pair of NVIDIA Tesla P100 GPUs.

\subsection{Results}

\begin{figure}[!t]
    \begin{center}
    \includegraphics[width=1\linewidth]{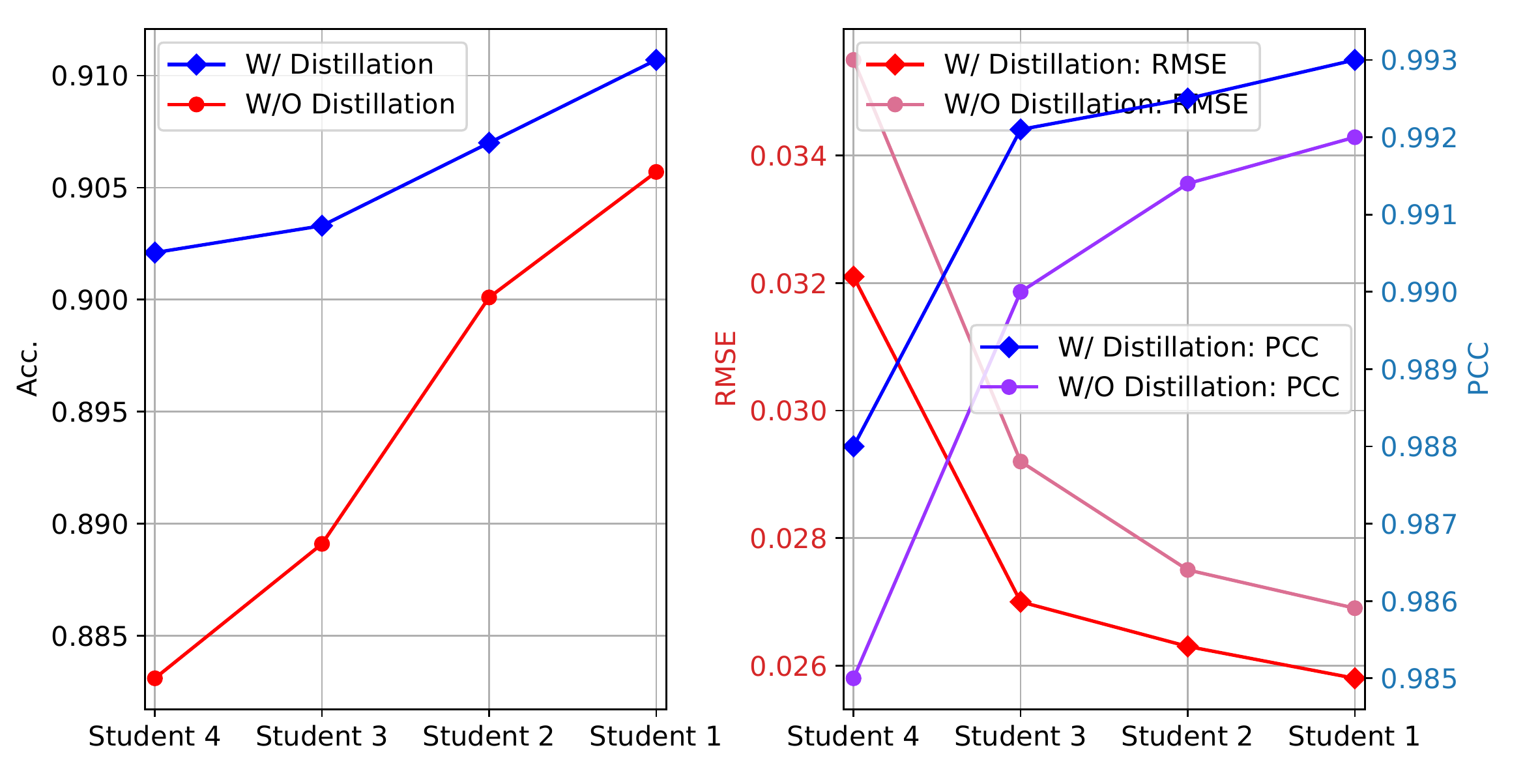} 
    \end{center}
\caption{The performance of the student with different model sizes for the SEED (\textbf{left}) and SEED-VIG (\textbf{right}) datasets, respectively. Comparison is performed between networks trained with distillation (labeled as 'W/ Distillation') and networks trained from scratch without distillation (labeled as 'W/O Distillation').}
\label{model_size}
\end{figure}

% Edited
\textbf{Student Model Size.}
We evaluate the impact of knowledge distillation on student networks with different number of parameters. To do so, we select several numbers of hidden units $M$ in the single LSTM layer, yielding $C\times (\sqrt M -2)^2$ number of lower level capsules for each of the four student networks. As shown in Table \ref{table:Distillation_Details}, each student network has a different number of parameters and respective compression ratio. Figure \ref{model_size} presents the performance of the students with different number of parameters. We observe consistent performance improvement for the student when our knowledge distillation framework is used, compared to when the student is trained from scratch in both datasets. For example, for the SEED dataset, the student $4$ achieves a $2\%$ boost with the help of the teacher, while for the SEED-VIG dataset, the performance of student $4$ improves by $0.0034$ in RMSE and $0.003$ in PCC.

% Edited
\textbf{Comparison with Existing Methods.}
We compare the performance of the best student, i.e. student 1, using our knowledge distillation framework with existing methods. Table \ref{table:SEED} shows the recent related work on the SEED dataset. Our student model $1$ with distillation obtains an accuracy of $0.9107\pm 0.0763$, achieving a good result. Table \ref{table:SEED-VIG} shows the related work on the SEED-VIG dataset. 
% Our student models $1$, $2$, and $3$ outperform the existing best results, with the help of knowledge distillation. Especially, 
Our student model obtains an RMSE of $0.0258\pm0.0095$ and a PCC of $0.9930\pm0.0047$, setting \textit{\textbf{new state-of-the-art results}}.

\begin{figure}[!t]
    \begin{center}
    \includegraphics[width=1\linewidth]{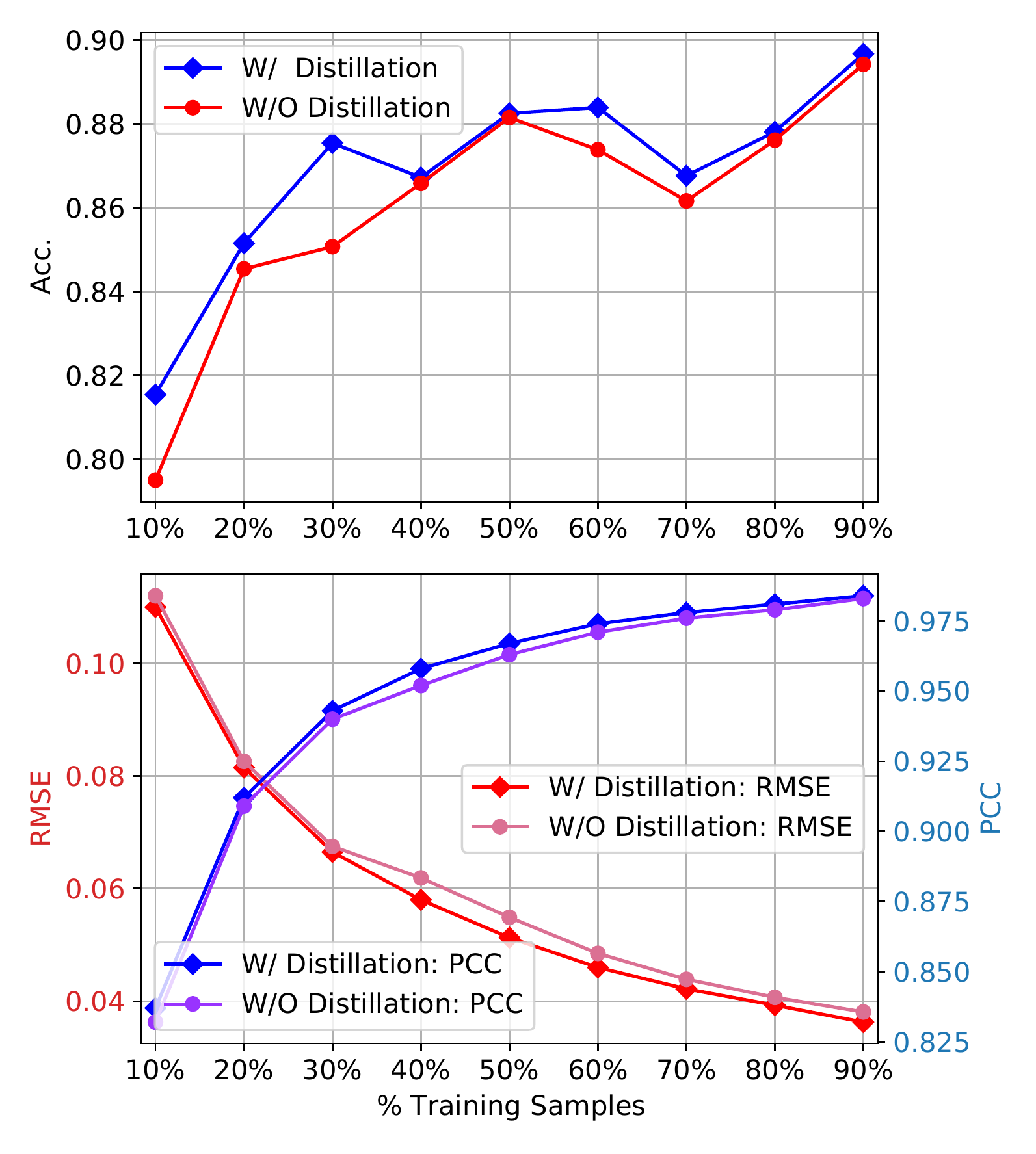} 
    \end{center}
\caption{The performance of the student network with different number of training samples for SEED (\textbf{top}) and SEED-VIG (\textbf{bottom}) datasets, respectively. Comparison is performed for networks trained with distillation (labeled as 'W/ Distillation') and networks trained from scratch without distillation (labeled as 'W/O Distillation').}
\label{few_training_samples}
\end{figure}

% Edited
\textbf{Fewer Training Samples.}
We investigate the role of distillation when fewer training samples are available. To do so, we randomly select different subsets of training samples ($10\%-90\%$) for training, with the same random seed throughout all experiments. We then compare the performance between the student trained without distillation and the student with the privileged information learned from the teacher, when different amount of training samples are used. We conduct the experiments using student model $4$ which has the smallest parameters among all students. Figure \ref{few_training_samples} shows the impact of our framework on the performance of the student network, when fewer training samples are available. For SEED, we observe convincing performance improvements brought by distillation when less than $40\%$ of training samples are available. Specifically, with only $30\%$ of training samples, the model performance improves by $3\%$ with the help of our proposed method. When more than $60\%$ of the training samples are used, improvements are marginal. Finally, the two models converge in the end. For SEED-VIG where the task is regression, utilizing a small subset of training samples may result in validation labels having output values not appearing during training. As a result, this experiment proves challenging when using very limited training samples.
% given that the task is regression, 
% unlike the classification task (SEED), 
% it is very challenging to perform regression label prediction with very limited training samples since training set may not contain the labels existing in test set. 
As shown in Figure \ref{few_training_samples}, the performance of both models significantly drops when the number of available training samples decreases. We observe that distillation doesn't help the student in performance when less than $40\%$ of training samples are available, while for larger subsets, slight improvements are consistently achieved. 
% This phenomenon is likely due to some labels in the test set are not included in such limited training samples. 
% However, distillation starts improving from , and consistent improvements of model with our distillation framework starting from $40\%$ to the end. 
% Overall, our knowledge distillation framework is capable of improving student network in both classification and regression tasks, when fewer training samples are available. 

% How to add this part to abstract and intro? This is also a contribution: sometimes, limited training samples are available for deployment  

\section{Conclusion}
We proposed a novel distillation framework on EEG representations for effective model compression. The framework is established on a capsule-based network and utilizes similarities among capsules for knowledge distillation. Our proposed method was applied on both classification and regression tasks on two popular public EEG datasets, in the field of affective computing. Our method shows strong performance for one dataset and achieves state-of-the-art for the other. Our experiments show that the improvement in performance of the student network with the teacher's privileged information compared to the same student trained without the teacher, increases as our student network is more compressed. Moreover, further experiments illustrate that our method is less sensitive to the size of the training set, helping the student in more effective learning when fewer training samples are available.

\label{sec:refs}

\bibliographystyle{IEEEtran}
\bibliography{IEEEabrv,refs}

\end{document}